\documentclass[letterpaper, 10 pt, conference]{ieeeconf}  
\IEEEoverridecommandlockouts                              
\usepackage{url}
\usepackage{amsmath,amssymb}
\usepackage{graphicx,color}
\usepackage{adjustbox}
\usepackage{booktabs}
\usepackage{multirow}
\usepackage{verbatim}
\usepackage{multicol}
\usepackage{subfig}
\usepackage{caption}
\usepackage{gensymb}
\usepackage{siunitx} 
\usepackage{pythonhighlight} 
\graphicspath{{figures/}}
\newcommand{\ba}{\mathbf{a}}

\newcommand{\bs}{\mathbf{s}}

\newcommand{\state}{\mathbf{s}}
\newcommand{\vmax}{v_{\rm max}}

\DeclareMathOperator*{\argmin}{arg\,min}
\DeclareMathOperator*{\minimize}{minimize}

\usepackage{pifont}
\newcommand{\remark}[3]{{\color{#2}[#1: #3]}}
\newcommand{\daniel}[1]{\remark{Daniel}{blue}{#1}}
\newcommand{\jonathan}[1]{\remark{Jonathan}{cyan}{#1}}

\newcommand{\Lawrence}[1]{\remark{Lawrence}{orange}{#1}}
\definecolor{britishracinggreen}{rgb}{0.23, 0.53, 0.19}

\newcommand{\jeffi}[1]{\remark{JI}{red}{#1}}

\definecolor{navy}{rgb}{0,0,0.5}

\usepackage[font={small}]{caption}
\def\tablename{Table}

\title{\LARGE \bf
Self-Supervised Learning of Dynamic \\ Planar Manipulation of Free-End Cables
}

\author{Jonathan Wang$^{1,*}$, Huang Huang$^{1,*}$, Vincent Lim$^{1}$, Harry Zhang$^{1}$,\\ Jeffrey Ichnowski$^1$, Daniel Seita$^1$, Yunliang Chen$^1$, Ken Goldberg$^1$
\thanks{*Equal contribution.}
\thanks{$^{1}$AUTOLAB at the University of California, Berkeley, USA.}%
}
\begin{document}

\maketitle
\thispagestyle{empty}
\pagestyle{empty}

\begin{abstract}
Dynamic manipulation of free-end cables has applications for cable management in homes, warehouses and manufacturing plants.
We present a supervised learning approach for dynamic manipulation of free-end cables, focusing on the problem of getting the cable endpoint to a designated target position, which may lie outside the reachable workspace of the robot end effector.
We present a simulator, tune it to closely match experiments with physical cables, and then collect training data for learning dynamic cable manipulation. We evaluate with 3 cables and a physical UR5 robot. Results over 32$\times$5 trials on 3 cables suggest that a physical UR5 robot can attain a median error distance ranging from 22$\%$ to 35$\%$ of the cable length among cables, outperforming an analytic baseline by 21$\%$ and a Gaussian Process baseline by 7$\%$ with lower interquartile range (IQR).
Supplementary material is available at \url{https://tinyurl.com/dyncable}.
\end{abstract}

\section{Introduction}\label{sec:intro}

Dynamic free-end cable manipulation is useful in a wide variety of cable management settings in homes and warehouses. For example, in a decluttering setting with cables, humans may dynamically manipulate a cable to a distant location to efficiently move it out of the way, cast it towards a human partner, or hang it over a hook or beam. Dynamic cable manipulation may also be advantageous when a cable must be passed through an obstacle to be retrieved on another side, requiring precision in the location of the cable endpoint. Casting a free-end cable radially outward from a base and subsequently pulling the cable toward the base --- which we refer to as ``polar casting'' and test in Section~\ref{ssection:methods} --- may be limited in free-end reachability, motivating the use of high-speed, dynamic motions. We use the generic term \emph{cable} to refer to a 1D deformable object with minimal stiffness, which includes ropes, wires, and threads. 

In this paper, we present a self-supervised learning procedure that, given the target location of a cable endpoint, generates a dynamic manipulation trajectory. Targets $(r, \theta)$ are defined in polar coordinates with radius $r$ and angle $\theta$ with respect to the robot base. We focus on planar dynamic cable manipulation tasks in which the cable lies on a surface and (after a reset action) a robot must dynamically adjust the cable via planar motions.
See Figure~\ref{fig:teaser} for an overview.
In prior work~\cite{harry_rope_2021}, we explored dynamic manipulation of fixed-end cables. In this work, we consider free-end cables, where the far end is unconstrained. 
We develop a simulation environment for learning dynamic manipulation of free-end cables in PyBullet~\cite{coumans2019}. We efficiently tune the simulator with differential tuning~\cite{de_tuning_2020} and train policies in simulation to get cable endpoints to reach desired targets. We perform physical experiments evaluating dynamic cable manipulation policies on a UR5 robot.
The contributions of this paper are:

\begin{figure}[t]
\center
\includegraphics[width=0.49\textwidth]{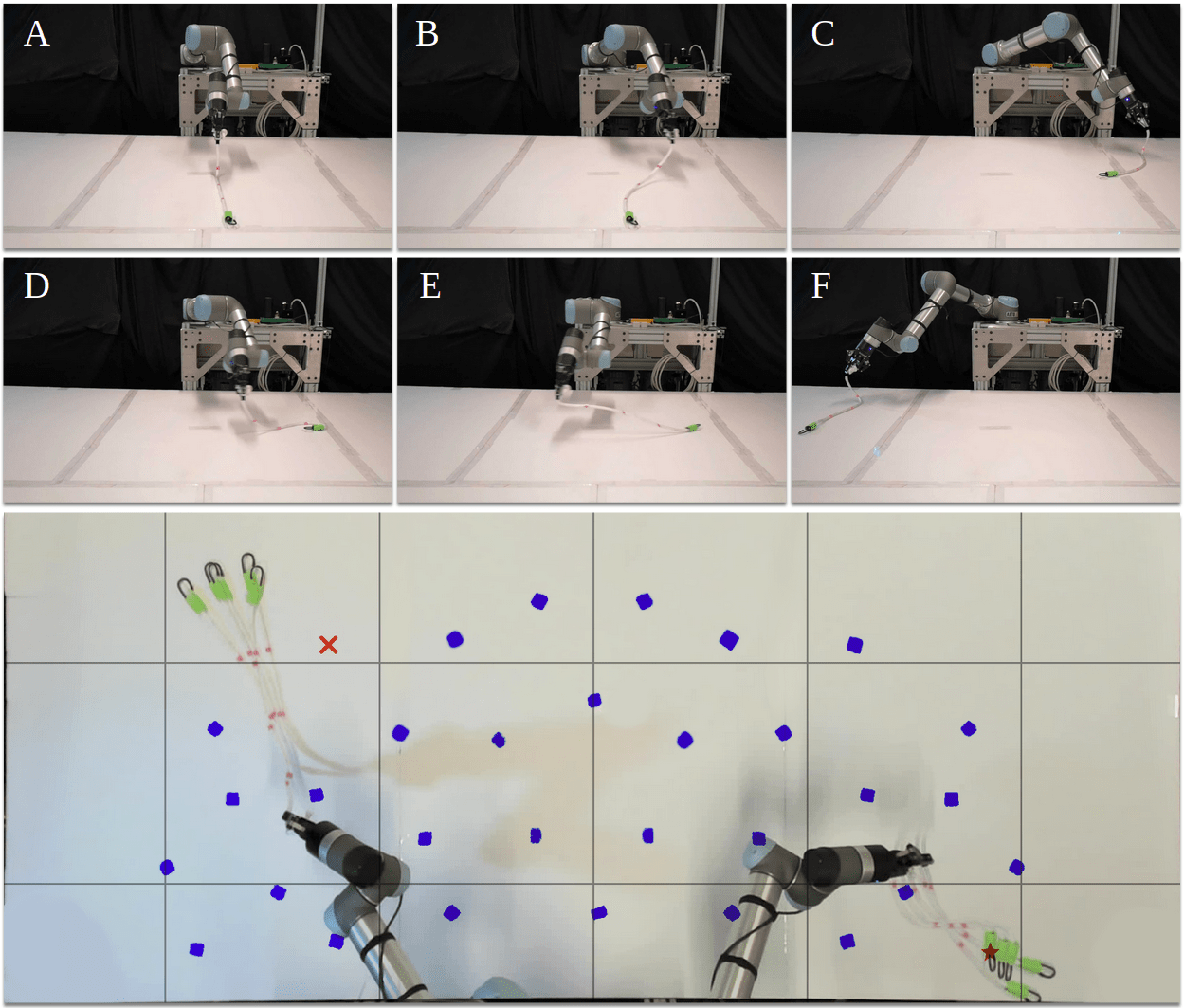}
\caption{
Examples of dynamic cable manipulation trajectories with a UR5 robot. After a reset (Section~\ref{ssec:resets}), the robot performs a dynamic planar action (Section~\ref{ssec:action}) with two waypoints. The top two rows show a frame-by-frame frontal view of the UR5 as it pulls towards one waypoint (A,B,C) and then towards the second (D,E,F) to get the cable endpoint to reach a desired target. The bottom image shows a top-down view of the setup. There are 32 target positions on the workspace plane marked with small tape, which we use to evaluate performance. The overlaid grid lines are spaced \SI{0.5}{\meter} apart. We additionally overlay a successful trial repeated 5 times, marked by the star target (lower right), and a failure trial repeated 5 times, marked by the red cross (upper left). 
}
\vspace{-16pt}
\label{fig:teaser}
\end{figure}

\begin{enumerate}
    \item Formulation of the dynamic planar manipulation of free-end cables problem.
    \item A dynamic simulator in PyBullet that can be tuned to accurately model dynamic planar cable manipulation.
    \item A dataset of 108,000 simulated trials and 1,120 physical trials with 3 cables and the UR5 robot.
    \item Physical experiments with a UR5 robot and 3 cables suggesting the presented method can achieve median error distance less than 35$\%$ of the cable lengths.
\end{enumerate}
\section{Related Work}\label{sec:rw}


\subsection{1D Deformable Object Manipulation}

Manipulation of 1D deformable objects such as cables has a long history in robotics. Some representative applications include surgical suturing~\cite{robot_heart_surgery_2006,suturing_autolab_2016, zhang2016health, zhang2021robots, zhang2020dex}, knot-tying~\cite{van_den_berg_2010,case_study_knots_1991,knot_planning_2003,tying_precisely_2016, sim2019personalization, elmquist2022art, avigal20206, avigal2021avplug, devgon2020orienting}, untangling ropes~\cite{rope_untangling_2013,grannen2020untangling,tactile_cable_2020, lim2021planar, lim2022real2sim2real}, deforming wires~\cite{wire_insertion_1997}, and inserting cables into obstacles~\cite{tight_tolerance_insertion_2015}. We refer the reader to Sanchez~et~al.~\cite{manip_deformable_survey_2018} for a survey. 

There has been a recent surge in using learning-based techniques for manipulating cables. A common setting is to employ pick-and-place actions with quasistatic dynamics so that the robot deforms the cable through a series of small adjustments, while allowing the cable to settle between actions. Under this paradigm, prior work has focused on cable manipulation using techniques such as learning from demonstrations~\cite{zeng_transporters_2020,seita_bags_2021, eisner2022flowbot3d, pan2022tax, pan2023tax, zhang2023flowbot++} and self-supervised learning~\cite{nair_rope_2017,ZSVI_2018}.
Prior work has also explored the use of quasistatic simulators~\cite{corl2020softgym} to train cable manipulation policies using reinforcement learning for eventual transfer to physical robots~\cite{lerrel_2020,yan_fabrics_latent_2020, yao2023apla, shen2024diffclip, jin2024multi}. In this work, we develop a fully dynamic simulator for dynamic cable manipulation to accelerate data collection for training policies, with transfer to a physical UR5 robot.
Other cable manipulation settings that do not necessarily assume quasistatic dynamics include sliding cables~\cite{one_hand_knotting_tactile_2007,zhu_sliding_cables_2019} or using tactile feedback~\cite{tactile_cable_2020} to inform policies. This paper focuses on dynamic planar actions to manipulate cables from one configuration to another.

\subsection{Dynamic Manipulation of Cables}

In dynamic manipulation settings, a robot takes actions rapidly to quickly move objects towards desired configurations, as exemplified by applications such as swinging items~\cite{swingbot_2020,diabolo_2020} or tossing objects to target bins~\cite{zeng_tossing_2019}. As with robot manipulation, these approaches tend to focus on dynamic manipulation of rigid objects.

A number of analytic physics models have been developed to describe the dynamics of moving cables, which can then be used for subsequent planning. For example,~\cite{fly_fishing_2004} present a two-dimensional dynamic casting model of fly fishing. They model the fly line as a long elastica and the fly rod as a flexible Euler-Bernoulli beam, and propose a system of differential equations to predict the movement of the fly line in space and time. 
In contrast to continuum models,~\cite{wang2011analysis} propose using a finite-element model to represent the fly line by a series of rigid cylinders that are connected by massless hinges. These works focus on developing mathematical models for cables, and do not test on physical robots.

In the context of robotic dynamic-cable manipulation,
Yamakawa~et~al.~\cite{dynamic_knotting_2010,yamakawa2012simple,high_speed_knotting_2013,one_hand_knotting_tactile_2007} demonstrate that a high-speed manipulator can effectively tie knots by dynamic snapping motions. They simplify the modelling of cable deformation by assuming each cable component follows the robot end-effector motion with constant time-delay. Kim and Shell~\cite{kim2016using} study a novel robot system with a flexible rope-like structure attached as a tail that can be used to strike objects of interest at high speed. This high-speed setting allows them to construct primitive motions that exploit the dynamics of the tail. They adopt the Rapidly-exploring Random Tree (RRT)~\cite{RRT_2001} and a particle-based representation to address the uncertainty of the object’s state transition. In contrast to these works, we aim to control cables for tasks in which we may be unable to rely on these simplifying assumptions.

Two closely related prior works are Zhang~et~al.~\cite{harry_rope_2021} and Zimmermann~et~al.~\cite{zimmermanndynamic}. Zhang~et~al.~\cite{harry_rope_2021} propose a self-supervised learning technique for dynamic manipulation of fixed-end cables for vaulting, knocking, and weaving tasks. They parameterize actions by computing a motion using a quadratic program and learning the apex point of the trajectory. In contrast to Zhang~et~al., we use \emph{free-end} cables. Zimmermann~et~al.~\cite{zimmermanndynamic} also study the dynamic manipulation of deformable objects in the free-end setting. They model elastic objects using the finite-element method~\cite{bathe2006finite} and use optimal control techniques for trajectory optimization. They study the task of whipping, which aims to find a trajectory such that the free end of the cable hits a predefined target with maximum speed, and laying a cloth on the table. Their simulated motions perform well for simple dynamical systems such as a pendulum, while
performance decreases for more complex systems like soft bodies due to the mismatch between simulation and the real world. We develop a learned, data-driven approach for robotic manipulation of free-end cables, focus on planar manipulation, and are interested in finding the best action such that the free end of the cable achieves a target position on the plane.

\section{Problem Statement}\label{sec:PS}

We consider a 6-DOF robot arm which controls a free-end cable with a small weight at the free end. The robot continually grips one end of the cable throughout each dynamic motion. We assume the cable endpoint lies on the plane and the robot end-effector takes actions at a fixed height slightly above the plane. The full configuration of the cable and the environment is infinite dimensional and complex to model, and we are interested in the cable endpoint. Thus, we concisely represent the state as ${\state=(r, \theta)}$, indicating the endpoint position on the plane in polar coordinates, with the robot base located at $(0, 0)$. The objective is to produce a policy $\pi$ that, when given the desired target ${\state^{(d)}=(r_d, \theta_d)}$ for the endpoint, outputs a dynamic arm action ${\ba=\pi(\state^{(d)})}$ that robustly achieves $\state^{(d)}$.
For notational convenience, we define $g_{\rm p2c} : \mathbb{R}^2 \to \mathbb{R}^2$ to be a function which converts a 2D point represented in polar coordinates to its 2D Cartesian coordinates. We also restrict targets $(r_d, \theta_d)$ such that they are reachable, that is, $r_d \leq r_{\rm max} + r_c$, where $r_{\rm max}$ is the maximal distance from the base to the robot wrist, and $r_c$ is cable length.

\begin{figure*}[t]
\center
\includegraphics[width=1.00\textwidth]{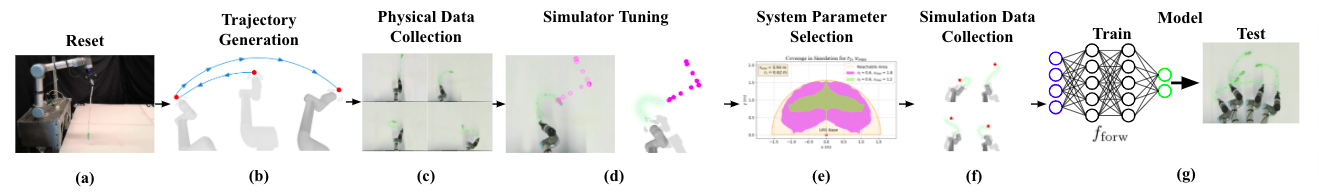}
\caption{
The process flow for learning dynamic planar cable manipulation (see Section~\ref{sec:method}). We find a reset motion (a) that consistently brings the cable to a fixed, reset state. We design and parameterize trajectories (b) for dynamically moving the cable endpoint, and execute these trajectories in real (c) to generate $\mathcal{D}_{\rm tune}$, which is then used to tune the dynamic simulator (d). We compare performance of system parameters to select those that maximize coverage and repeatability (e), and use these fixed parameters to generate $\mathcal{D}_{\rm sim}$ (f) and $ \mathcal{D}_{\rm real}$. We learn $f_{\rm forw}$ using $\mathcal{D}_{\rm sim}$ along with fine-tuning on $\mathcal{D}_{\rm real}$. The $f_{\rm forw}$ is evaluated on real targets with a physical robot.
}
\vspace*{-10pt}
\label{fig:process}
\end{figure*}

\section{Learning Framework}\label{sec:method}

Variations in cable properties such as mass, stiffness, and friction, mean that a robot may require different trajectories to get endpoints to reach the same target. Directly estimating cable configurations after a motion is challenging due to complex dynamics and the difficulty of state estimation.
In this work, we use the following pipeline for manipulating cable endpoints towards desired targets, summarized in Figure~\ref{fig:process}. To reduce uncertainty in state estimation, we start each motion from a consistent initial state. We do this with a \emph{reset procedure} with high \emph{repeatabilty} (Section~\ref{sec:resets-repeat}). 
Next, for a given cable, we collect video observations of cable motions that start from the reset state (Section~\ref{ssec:data-gen}). We use this dataset to automate the tuning of a cable simulator to match the behavior of a physical cable (Section~\ref{ssec:simulator}).
From simulated data, we then train a neural network model that maps trajectories to predicted endpoints that the policy $\pi$ plans through to output a trajectory that achieves the desired target (Section~\ref{ssec:supervised}).

\subsection{Manual Reset and Repeat to Measure Repeatability}\label{sec:resets-repeat}

A prerequisite for this problem is that actions corresponding to state transitions must be repeatable. We design and parameterize trajectories with the goal of maximizing repeatability, which we measure by executing the same $\ba$ multiple times, and recording the variance in the final endpoint location.
To get the cable into a \emph{reset} state, the robot performs a sequence of trajectories that reliably places the cable into a consistent state. We find an open-loop sequence that includes lifting the cable off a surface and letting it hang until it comes to rest. This may be part of a reliable reset method, as it undoes torsion consistently with sufficient time. For further details, see Section~\ref{ssec:resets} and Figure~\ref{fig:reset}. 

\subsection{Dataset Generation}\label{ssec:data-gen}

To facilitate training, we take inspiration from Hindsight Experience Replay (HER)~\cite{HER_2017} by executing an action $\ba$, recording the resulting endpoint location $(r_d, \theta_d)$, and assuming that it was the intended target. We store each transition $(\ba, \bs)$ in the dataset. We create three datasets for the cable: $\mathcal{D}_{\rm tune}$, real trajectories for tuning the simulator (Section~\ref{ssec:diff-ev}), $\mathcal{D}_{\rm sim}$, trajectories executed in the tuned simulator to train a forward dynamics model (Section~\ref{ssec:supervised}), and $\mathcal{D}_{\rm real}$, real trajectories used for fine-tuning the model trained on simulation examples. For notational convenience, we define a model $f$ trained on $\mathcal{D}_{\rm sim} + \mathcal{D}_{\rm real}$ as a model pre-trained on $\mathcal{D}_{\rm sim}$ and later fine-tuned with $\mathcal{D}_{\rm real}$. As discussed in Section~\ref{ssec:data-collect}, $\mathcal{D}_{\rm tune}$ and $\mathcal{D}_{\rm real}$ are both from real trajectories, but may have been generated from different system parameters.

\subsection{Cable Simulator}
\label{ssec:simulator}

  
We create a dynamic simulation environment to efficiently collect data for policy training. Using the PyBullet~\cite{coumans2019} physics engine, we model cables as a string of capsule-shaped rigid bodies held together by 6-DOF spring constraints, as the adjustable angular stiffness values of these constraints can model twist and bend stiffness of cables. The environment also includes a flat plane as the worksurface, and a robot with the same IK as the physical robot. We tune 10 cable and worksurface parameters to best capture cable physics: twist stiffness, bend stiffness, mass, lateral friction, spinning friction, rolling friction, endpoint mass, linear damping, angular damping, and worksurface friction, and tune the parameters (Section~\ref{ssec:diff-ev}).




\subsection{Training the Trajectory Generation Model} 
\label{ssec:supervised}


We generate trajectories via a learned forward dynamics model. The policy $\pi$ first samples multiple input actions $\ba$ to form a set $\mathcal{A}$ of candidate actions. It then passes each action through the forward dynamics model $f_{\rm forw}$, which outputs the predicted endpoint location $(\hat{x}, \hat{y})$ in Cartesian space. Given a target endpoint $\bs$ (in polar coordinates), the action ${\ba = \pi(\bs)}$ selected is the one minimizing the Euclidean distance between the predicted and target endpoints:
\begin{equation}\label{eq:finv}
\pi(\bs) = \argmin_{\ba \in \mathcal{A}} \; \| f_{\rm forw}(\ba) - g_{\rm p2c}(\bs) \|_2
\end{equation}
where both $f_{\rm forw}(\ba)$ and $g_{\rm p2c}(\bs)$ are expressed in Cartesian (not polar) coordinates. We generate a dataset $\mathcal{D}_{\rm sim}$ for the cable via grid-sampled actions using the cable simulator with tuned parameters to match real, and then use it to train $f_{\rm forw}$.

\section{Evaluation}
\label{sec:experiments}

\begin{figure}[t]
  \center
  \includegraphics[width=0.38\textwidth]{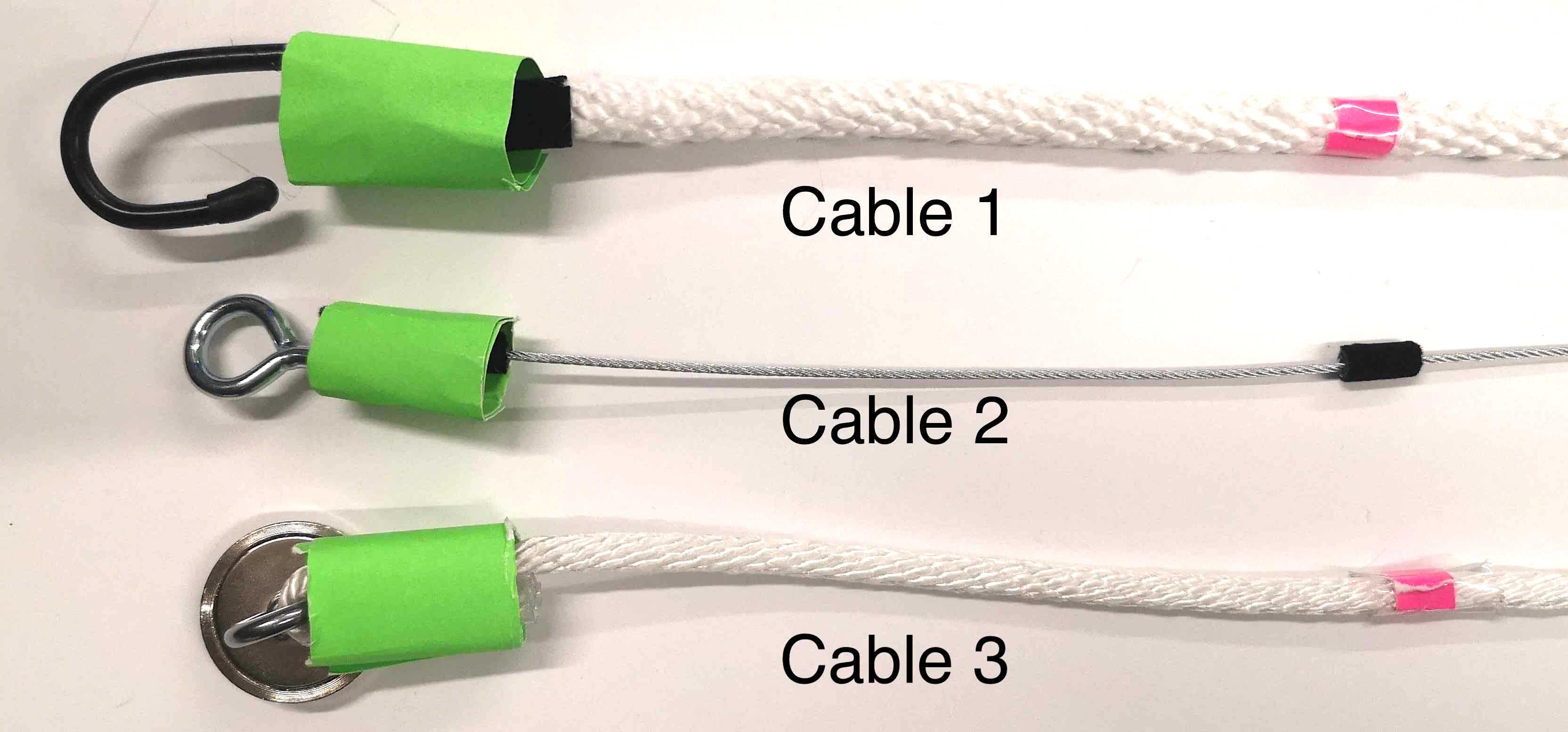}
  \caption{
  Three cables used in physical experiments, with two polyester ropes and one braided iron wire. Each endpoint has an attached mass, either hooks (first two) or a magnet (third cable). From top to bottom, the cable lengths are \SI{0.62}{\meter}, \SI{0.65}{\meter}, and \SI{0.67}{\meter} without the attachment, and \SI{0.67}{\meter}, \SI{0.68}{\meter}, and \SI{0.68}{\meter} with the attachment. The diameters are \SI{0.011}{\meter}, \SI{0.002}{\meter}, and \SI{0.007}{\meter}.
  }
  \vspace*{-10pt}
  \label{fig:cables}
\end{figure}

We instantiate and evaluate the presented method from Section~\ref{sec:method} on a physical UR5 robot (Figure~\ref{fig:teaser}) with three cables of different properties (Figure~\ref{fig:cables}). We use a table that is \SI{2.75}{\meter} wide and extends \SI{1.50}{\meter} from the base of the robot. Each cable has a weighted attachment, either a hook or a magnet to resemble a home or industrial setting in which the passing of the cable sets up a downstream task such as plugging into a mating part. The table is covered with foam boards for protection and to create a consistent friction coefficient. To facilitate data collection, we attach a green markers to the cables. We record observations from an overhead Logitech Brio 4K webcam recording 1920$\times$1080 video at 60 frames per second.


\subsection{Reset}\label{ssec:resets}

\begin{figure}[t]
\center
\includegraphics[width=0.49\textwidth]{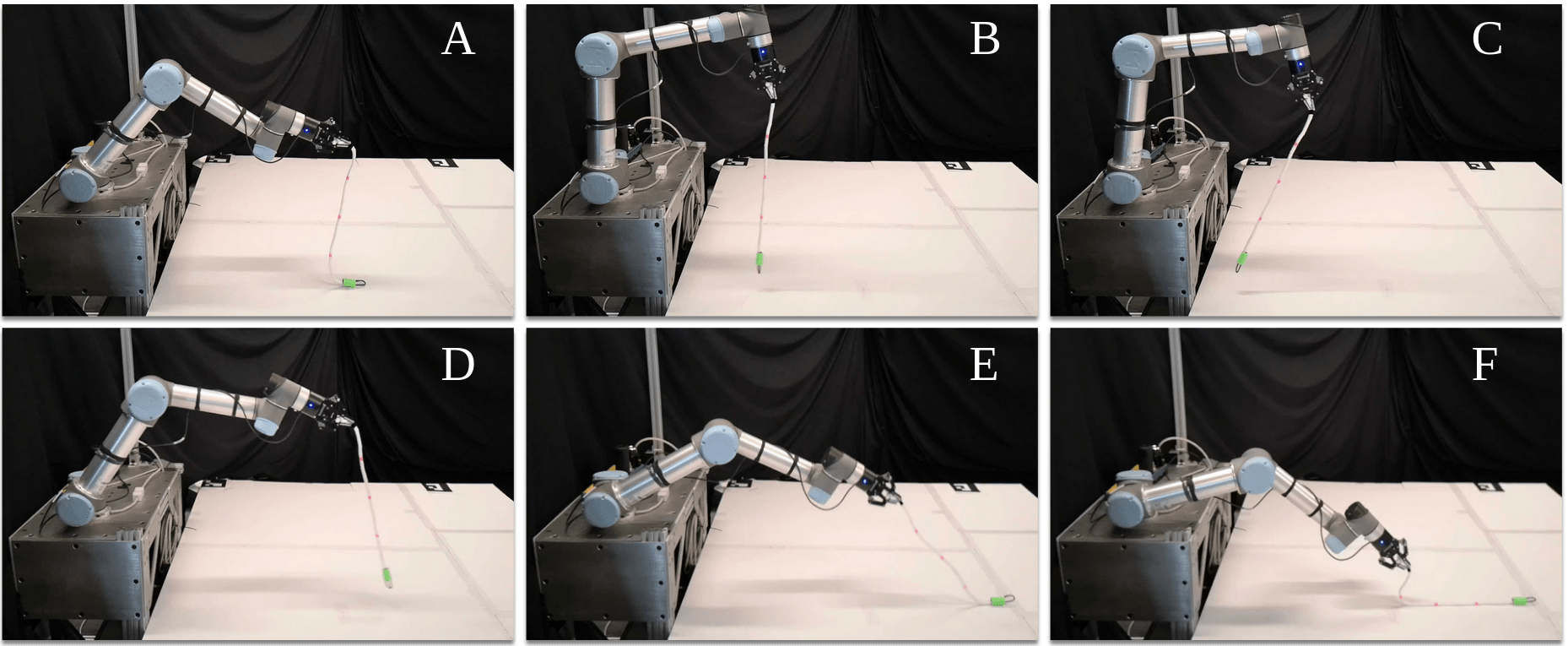}
\caption{
Overview of the reset procedure with the UR5, which occurs at the start of each dynamic manipulation. The photos show a frame-by-frame side view of the UR5 as it resets by letting the cable settle (A,B), pulling back (C), casting forward (D,E), and dragging back slightly (F). See Section~\ref{ssec:resets} for more details. After this reset, the robot executes planar actions, as described in Section~\ref{ssec:action}.
}
\vspace*{-15pt}
\label{fig:reset}
\end{figure}

We define the following reset procedure:
\begin{enumerate}
\item Lift the cable vertically up with the free-end touching the worksurface to prevent the cable from dangling.
\item Continue to lift the cable up such that the free-end loses contact with the worksurface.
\item Let the cable hang for 3 seconds to stabilize.
\item Perform polar casting to swing the free-end forward to land at $(r_d, \theta_d=0)$, then slowly pull the cable towards the robot until the reset position is reached.
\end{enumerate}
Figure~\ref{fig:reset} shows a frame-by-frame overview of the reset procedure, which the UR5 robot executes before each dynamic planar cable manipulation action (see Section~\ref{ssec:action}) to ensure the same initial cable configuration.

\subsection{Planar Action Variables}\label{ssec:action}
For dynamic manipulation, the robot must take an action that induces the cable to swing to reach the target.  We desire several action properties: (1) it should be a quick non-stop and smooth motion, (2) it should require a low-dimensional input that facilitates data generation and training, (3) it has a subset of parameters that can be fixed yet still allow the varying parameters to generate motions that will carry out the task, and (4) it should be interpretable.  We hand craft a function to have these properties based on observing a human attempting the same task, and observe that most motions arc one direction, then arc back, before coming to a stop. 

\begin{figure}[t]
\center
\includegraphics[width=0.4\textwidth]{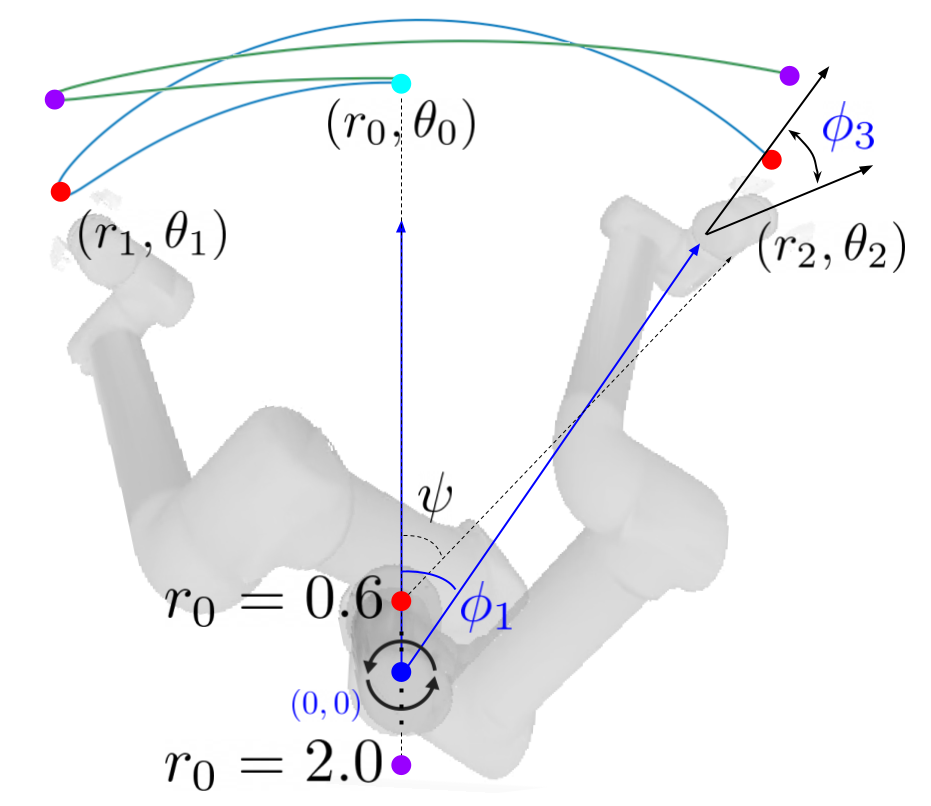}
\caption{
Example of two sets of splines that are traced by the end effector during the trajectory (Equation \ref{eq:act2_wrist}) of the UR5, for two different $r_0$ values: $r_0 \in \{0.6, 2.0\}$ (not drawn to scale). The reset procedure brings the end effector to $(r_0, \theta_0)$ in the polar space before the start of each dynamic motion. Decreasing $r_0$ increases the curvature of the trajectory. 
}
\vspace*{-10pt}
\label{fig:spline}
\end{figure}

To design the trajectory function, we define a coordinate frame and set of coordinates that the trajectory function uses to define an action (Figure~\ref{fig:spline}).
An offset $r_0$ from the reset position defines the origin of a polar coordinate frame.  
In this coordinate frame, we define 3 polar coordinates that define the motion.  The motion starts at $(r_0, \theta_0)$ then interpolates through $(r_1, \theta_1)$ and comes to a rest at $(r_2, \theta_2)$.
We use an inverse kinematic (IK) solver to compute the robot configuration through the motion.  We observe that when polar coordinate origin aligns with the base of the robot, the angular coefficient corresponds 1:1 with the base rotation angle $\phi_1$, and the IK solution for that joint is trivial.
To increase the sweeping range of the robot, we also include a wrist joint rotation $\phi_3$ in the trajectory function.

The handcrafted trajectory function parameterizes each action by a fixed set of \emph{system parameters} and a variable set of \emph{action variables}.
The system parameters are: the origin offset $r_0$ and the maximum linear velocity $\vmax$.
Increasing $r_0$ creates a wider circle, thus flattening the arc.  In Figure~\ref{fig:spline}, the smaller value of $r_0$ results in the blue arc, while the larger value of $r_0$ results in the flatter green arc. Once we determine the system parameters (Section~\ref{ssec:sys-param}), they remain fixed, and only the action variables vary during training.

We consider two sets of action variables, $A_1$ and $A_2$. The set $A_1$ consists of three values:
\begin{equation}\label{eq:act2}
    \ba = (\theta_1, \theta_2, r_2),
\end{equation}
while the set $A_2$ consists of four:
\begin{equation}\label{eq:act2_wrist}
    \ba = (\theta_1, \theta_2, r_2, \psi).
\end{equation}
The first three parameters correspond to coefficients of polar coordinates that define the motion (Fig.~\ref{fig:spline}).  The fourth is the wrist joint's rotation about the $z$-axis.  We convert these coordinates to a trajectory in polar coordinates by using a cubic spline to smoothly interpolate the radial coefficient from $r_0$ to $r_2$, and by using a maximum-velocity spline to interpolate the angular coefficient from $\theta_0$ to $\theta_1$ and from $\theta_1$ to $\theta_2$, with maximum velocity $\vmax$.  We implement the maximum-velocity spline with a jerk-limited bang-bang control, having observed that the UR5 has difficulty following trajectories with high jerk~\cite{ichnowski2020djgomp}. We assume a direction change between the two angular splines, and thus the angular velocity and acceleration at $\theta_1$ is 0.  Once we have the polar coordinate trajectory of the end effector, we convert it to a trajectory in joint space using an IK solver by discretizing the trajectory at the time interval of the robot's control cycle.

We observe the coverage space for $A_1$ is limited in practice (Figure~\ref{fig:coverage}), and thus consider the action set $A_2$. 
When using $A_2$, the wrist rotates $\psi - \theta_2$ radians with a maximum velocity spline rotation. In an ablation study in Section~\ref{ssec:system-ablation}, we examine the impact of the wrist angle parameter.

This parameterization allows us to take advantage of the symmetrical aspect of the problem. For all datasets, we sample actions such that $\theta_1 < 0$, $\theta_2 > 0$, and $\psi \geq \theta_2$, to obtain targets on the right of the workspace axis of symmetry. If $(\theta_1, \theta_2, r_2, \psi)$ produces target $(r_d, \theta_d)$, $(-\theta_1, -\theta_2, r_2, -\psi)$ will produce $(r_d, -\theta_d)$. Therefore, during evaluation, for targets $(r_d, \theta_d)$ on the left of the axis, we take $\pi((r_d, -\theta_d)) = (\hat{\theta}_1, \hat{\theta}_2, \hat{r}_2, \hat{\psi})$ and output $(-\hat{\theta}_1, -\hat{\theta}_2, \hat{r}_2, -\hat{\psi})$.


\subsection{Self-Supervised Physical Data Collection}\label{ssec:phys-data}
To create $\mathcal{D}_{\rm tune}$, we generate and execute physical trajectories $(\theta_1, \theta_2, r_2, \psi)$, uniformly sampling each parameter such that the trajectory does not violate joint limits.
The execution for each trajectory is about \SI{20}{\second}, including \SI{15}{\second} for the reset motion and \SI{5}{\second} for the planar motion, thus motivating the need for more efficient self-supervised data generation in simulation. Videos for each planar motion are recorded. We collect physical trajectories for tuning the simulator and fine tuning the model.

\subsection{Simulator Tuning}\label{ssec:diff-ev}
Instead of using grid search or random search,
we search for parameters using Differential Evolution (DE)~\cite{de_tuning_2020}, optimizing for a similarity metric. DE does not require a differentiable optimization metric, and scales reasonably with a large number of parameters compared to grid search and Bayesian optimization~\cite{de_tuning_2020}.

We denote the collective set of PyBullet simulation parameters as $\theta_{\rm sim}$. For each trajectory $m_j$ in the training set $\mathcal{D}_{\rm tune}$ of $M$ total \emph{physical} trajectories, we record the 2D location of the cable endpoint, $p_t = g_{\rm p2c}(\bs_{\rm real}) = (x_t, y_t)$, at each time step $t$ spaced \SI{100}{\milli\second} apart (see Figure~\ref{fig:process}(d)). The tuning objective is to find parameters $\theta_{\rm sim}$ that minimize the average $L^2$ distance between the cable endpoint in simulation, $\hat{p}_t = g_{\rm p2c}(\bs_{\rm sim}) = (\hat{x}_t, \hat{y}_t)$, to the real waypoint at each $t$:
\begin{equation}\label{eq:tuning}
    \minimize_{\theta_{\rm sim}} \;\; \epsilon_{\rm trajs} = {\frac{1}{M}\sum_{j}^M{\frac{1}{N_j}\sum_i^{N_j}{\lVert p_i - \hat{p}_i \rVert}_2}}.
\end{equation}
The number of waypoints compared for each trajectory is $N_j = \lfloor {T_j}/{100} \rfloor$, where $T_j$ is the duration of $m_j$ in milliseconds. For each cable, we collect 60 physical trajectories and evaluate on an additional 20 trajectories where $(\theta_1, \theta_2, r_2, \psi)$, for each trajectory are uniformly sampled. Increasing the number of trajectories to 80 leads to an increase in $\epsilon_{\rm trajs}$ by $2\% - 12\%$, suggesting that additional trajectories would not improve tuning.

\begin{table}[t]
  \setlength\tabcolsep{5.0pt}
  \centering
  \caption{
    Results for Differential Evolution (Section~\ref{ssec:diff-ev}) on a set of 20 validation trajectories for 3 cables (see Figure~\ref{fig:cables}). The two metrics are the median final $L^2$-distance, which is the 2D Euclidean distance between endpoint locations in simulation and reality after a trajectory terminates, and $\epsilon_{\rm trajs}$, defined in Equation $\ref{eq:tuning}$. We compare the performance of parameters found by DE with that of PyBullet's default parameter settings. Values are expressed in percentages of cable length.
  }
  \small
  \begin{tabular}{@{}lllll}
  \toprule
  Metric & Cable 1  & Cable 2  & Cable 3  \\
  \midrule
  (Default) Median final $L^2$-distance & 35.0\% & 25.8\% & 24.5\% \\
  (Default) $\epsilon_{\rm trajs}$ & 38.2\% & 29.4\% & 24.5\% \\
  (DE) Median final $L^2$-distance & \textbf{14.7}\%  & \textbf{12.2}\% & \textbf{14.5}\% \\
  (DE) $\epsilon_{\rm trajs}$ & \textbf{28.4}\% & \textbf{23.4}\% & \textbf{18.7}\% \\
  \toprule
  \end{tabular}
  \vspace*{-10pt}
  \label{tab:tuning}
\end{table}  


Table~\ref{tab:tuning} compares similarity between cable behavior in simulation and real. For each cable, DE was able to reduce the discrepancy in the endpoint location throughout and after the trajectory, compared to PyBullet's default parameter settings. While $50\%$ of trajectories have final $L^2$-distance less than $15\%$, evaluation results suggest a reality gap still remains for certain trajectories, motivating the use of fine-tuning as discussed in Section~\ref{ssec:data-collect}.

\subsection{System Parameter Selection}\label{ssec:sys-param}

\begin{figure}[t]
\centering
\subfloat{\includegraphics[width=0.39\textwidth]{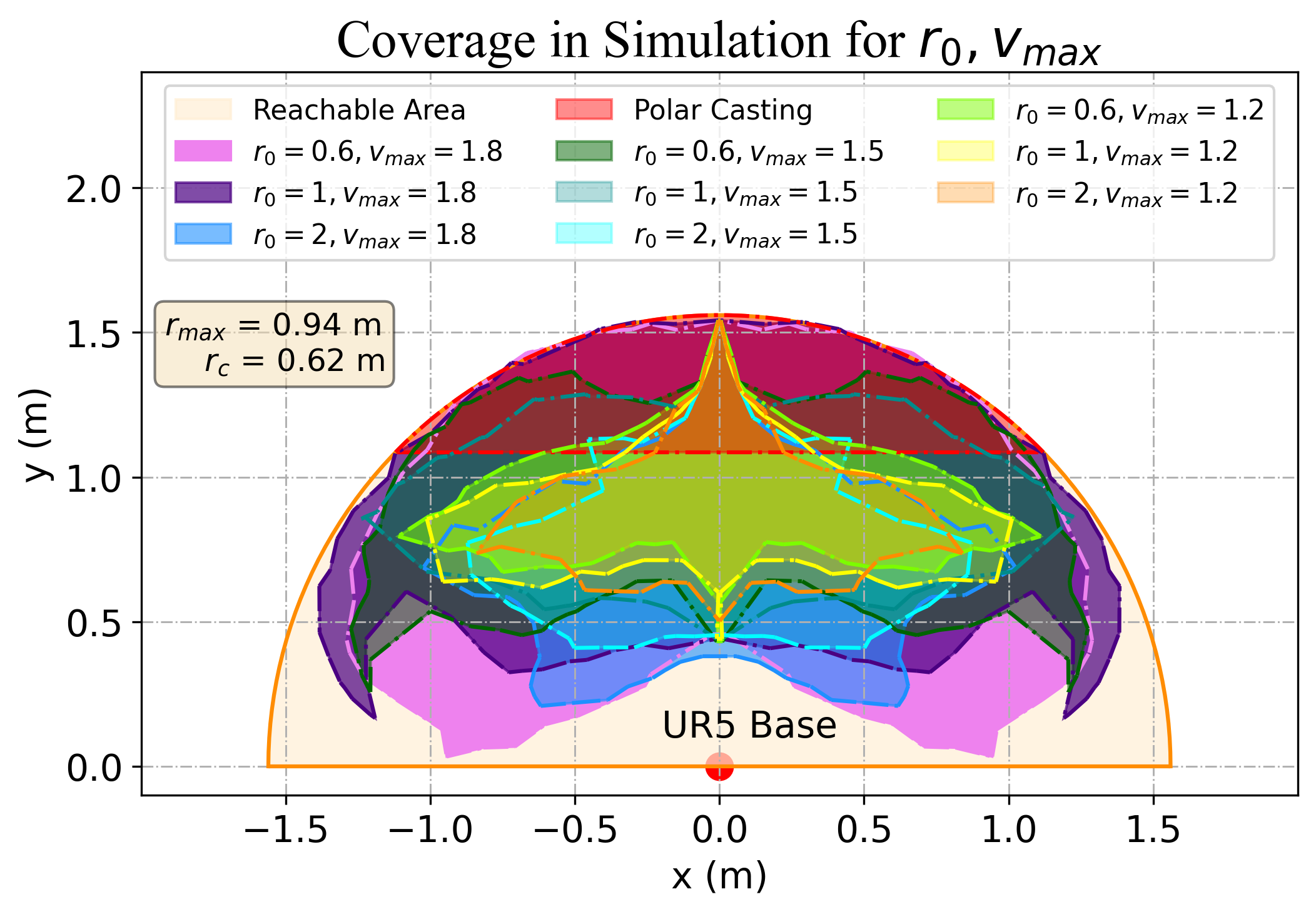}}

\subfloat{\includegraphics[width=0.39\textwidth]{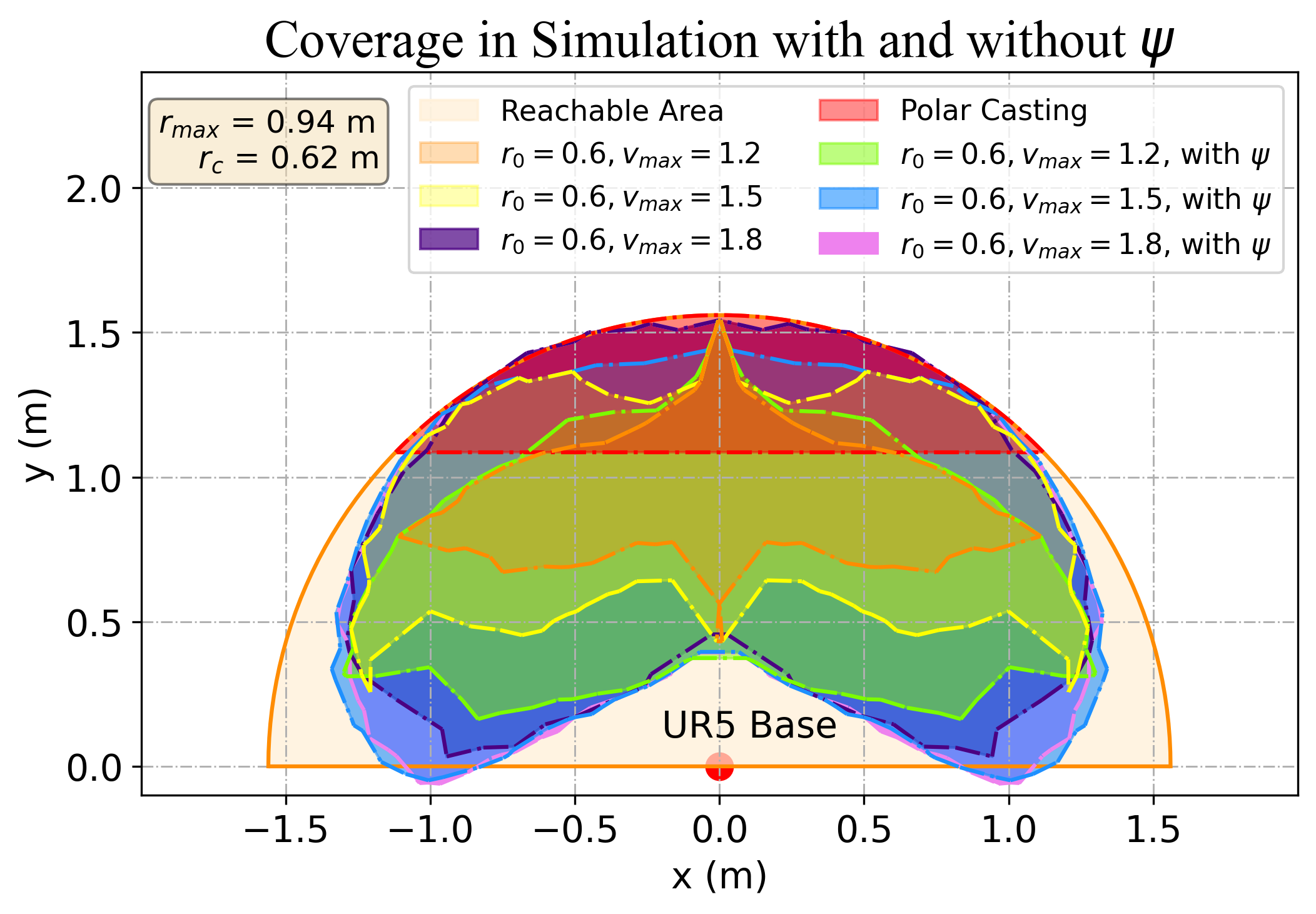}}
\caption{
Plots showing the area on the workspace plane that is reachable by the cable endpoint across sampled trajectories for each system parameter combination in simulation, with simulation parameters tuned for Cable 1. The reachable area of the endpoint is defined by a semicircle with radius $r_{\rm max} + r_c$. Each plot is generated by executing grid sampled actions $(\theta_1, \theta_2, r_2)$ with $\theta_1 < 0$ and $\theta_2 > 0$ in the simulator (and $\psi \geq \theta_2$ if with $\psi$), reflecting endpoints across the vertical axis due to symmetrical actions, and applying a concave hull to the coordinates of the endpoints. All planar actions can achieve coverage beyond the area reachable by polar casting.
}
\label{fig:coverage}
\vspace*{-10pt}
\end{figure}

To find the optimal $(r_0, \vmax)$ combination that maximizes endpoint coverage and repeatability, we search in simulation and real. For all grid samples, we filter actions that exceed joint or velocity limits. We use ${r_0 \in \{0.6, 1.0, 2.0 \}}$ and $\vmax \in \{1.2,1.5,1.8\}$ for $3\times3=9$ combinations. For each $r_0$ and $\vmax$ pair, we generate actions $\ba=(\theta_1,\theta_2,r_2)$ via grid sampling. We generate $15\times15\times15$ actions with $15$ values for each parameter and execute them in simulation. We evaluate the endpoint coverage for each $(r_0, \vmax)$ combination by aggregating all resulting endpoints from these actions. Fig.~\ref{fig:coverage} compares endpoint coverage among different $(r_0, \vmax)$ combinations, overlaid on top of a semicircle with radius $r_{\rm max} + r_c$, representing the maximum possible coverage.

Higher velocity actions tend to give higher coverage, but also higher uncertainty, resulting in less repeatability. To evaluate the repeatability, we execute each sampled action 5 times in the physical environment and calculate the standard deviation of the distance between the cable endpoint position to the mean position. For trials where the cable endpoint leaves the table (i.e., ``off-table''), we remove these trials from the standard deviation calculation. To evaluate repeatability for a given $r_0$ and $\vmax$, we average the standard deviations across actions generated via grid sampling. We seek an $r_0$ and $\vmax$ combination that maintains repeatability with high coverage.
%

With the $r_0$ that maximizes coverage, we evaluate the effect of the additional wrist rotation parameter $\psi$ on coverage and repeatability. 
For $\vmax \in \{1.2,1.5,1.8\}$, we generate $15\times15\times15$ actions for $a=(\theta_1,\theta_2,r_2)$ and $10\times10\times10\times5$ actions for $a=(\theta_1,\theta_2,r_2, \psi)$ and execute them in simulation to evaluate coverage. In real, we generate and execute $5\times5\times5$ actions for $a=(\theta_1,\theta_2,r_2)$ with $\vmax \in \{1.2,1.5\}$ and $3\times3\times3\times3$ actions for $a=(\theta_1,\theta_2,r_2, \psi)$ with $\vmax \in \{1.5,1.8\}$ to evaluate repeatability.

\subsection{Training Data Collection and Model Training}\label{ssec:data-collect}

After selecting the most ideal system parameters for coverage and repeatability, we use actions generated with these fixed values to learn a forward dynamics model as in Section~\ref{ssec:supervised}. To generate training data $\mathcal{D}_{\rm sim}$, we grid sample $\theta_1$, $\theta_2$, $r_2$ and $\psi$, and generate 36,000 $(\ba, \bs)$ simulated transitions to train $f_{\rm forw}$ to predict $\bs$ given $\ba$. We additionally execute 200 grid-sampled actions and record the transitions in the physical setup to generate $\mathcal{D}_{\rm real}$, which is used for fine-tuning the dynamics model with real data.


We use a feedforward neural network for $f_{\rm forw}$, with 4 fully connected layers with 256 hidden units each. The model is trained to minimize the $L^2$ error on data from $\mathcal{D}_{\rm sim}$:
\begin{equation}\label{eq:forw}
    \sum_{(\ba, \bs) \in \mathcal{D}_{\rm sim}} {\lVert f_{\rm forw}(\ba) - g_{\rm p2c}(\bs) \rVert}_2.
\end{equation}
During evaluation, we sample 50,000 candidate actions using a grid sample on the joint-limit-abiding action variables, and select the action that minimizes the $L^2$ distance between the predicted and desired endpoints (see Equation~\ref{eq:finv}).



\begin{figure}[!tbp]
    \centering
    \includegraphics[width=0.49\columnwidth]{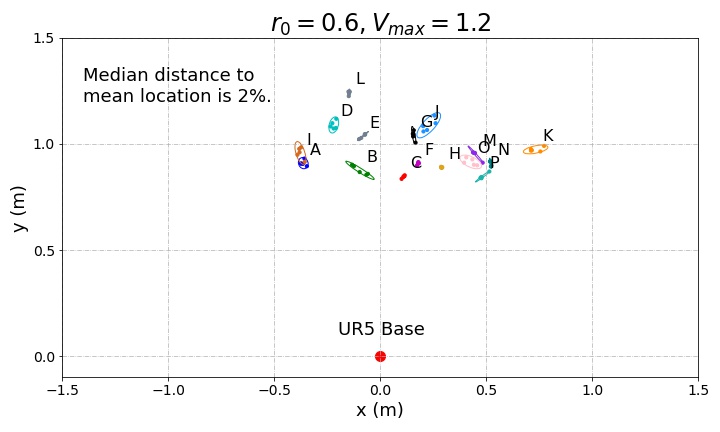}
    \includegraphics[width=0.49\columnwidth]{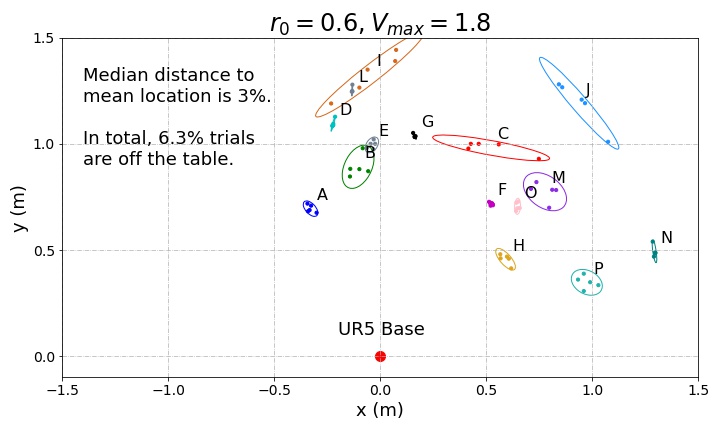}
    \includegraphics[width=0.49\columnwidth]{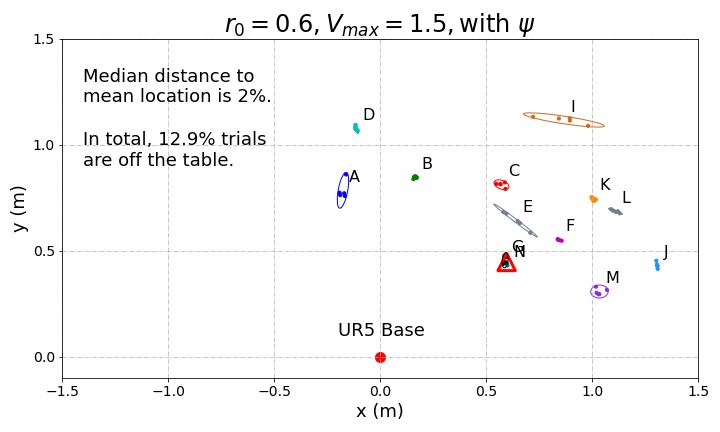}
    \includegraphics[width=0.49\columnwidth]{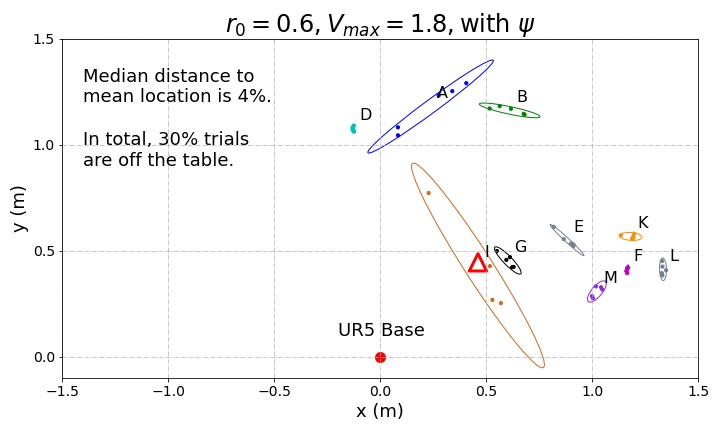}
    \caption{
    Repeatability analysis for Cable 1. The points of each color correspond to the locations of the 5 trials with the same action. A 95\% confidence ellipse is fit for each cluster. A small red triangle around the number indicates that there is at least one off-table trial for that action (see Sec.~\ref{sec:results}). Lower $\vmax$ actions without $\psi$ are shown to be more repeatable than higher $\vmax$ actions with $\psi$.
}
\label{fig:repeat}
\vspace*{-10pt}
\end{figure}

\begin{figure*}[!h]
    \centering
    \includegraphics[width=\textwidth]{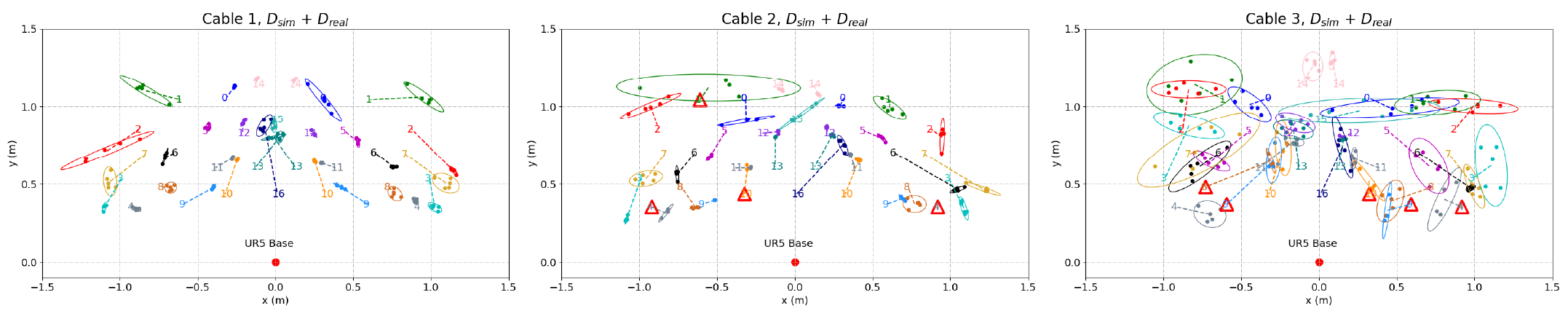}
    \caption{
    Evaluation results of the forward dynamics models trained with $\mathcal{D}_{\rm sim} + \mathcal{D}_{\rm real}$ for each cable. The number represents each target and the point represents each endpoint corresponding to the target with the same color. The ellipse is the 95\% confidence ellipse and the dashed line connects the center of the ellipse to the target representing the mean error distance. A red triangle indicates there is at least one off-table trial or occluded trial for this target. There are 7 off-table trials for Cable 2 and 1 off-table trial for Cable 3. 
}
\label{fig:repeat_multi_cable}
\vspace*{-10pt}
\end{figure*}

\section{Results}\label{sec:results}

We extract the endpoint position of the cable from an overhead image after each executed action. For off-table cases and occluded case where the endpoint is occluded by the robot arm, we exclude them from the error analysis. We'll address the avoidance of these cases in the future work.

\subsection{Methods}\label{ssection:methods}

We benchmark performance of the forward dynamics model trained on (1) $\mathcal{D}_{\rm real}$ only, (2), $\mathcal{D}_{\rm sim}$ only, and (3) $\mathcal{D}_{\rm sim}$ first, then fine-tuned on $\mathcal{D}_{\rm real}$. See Section~\ref{ssec:data-gen} for dataset details. Along with these three forward models, we test with two baseline methods, one analytic and one learned:

\textbf{Polar Casting.} Given a target cable endpoint location $(r_d, \theta_d)$, first perform a casting motion that causes the free endpoint to land at $(r_{\rm max} + r_c, \theta)$. To perform this motion, the robot rotates $\theta_d$ radians, and executes a predefined casting motion. After the casting motion, the robot slowly pulls the cable toward the base for a distance of $r_{\rm max} + r_c - r_d$. Targets $(r_d, \theta_d)$ achievable by this method are limited to the circular segment seen in Figure~\ref{fig:coverage}. In Cartesian coordinates, the arm is limited to a minimum $y$ coordinate of $\SI{0.466}{\meter}+r_c$ to prevent the end effector from hitting the robot's supporting table, where $\SI{0.466}{\meter}$ is the distance from the center of the robot base to the edge of the table. 


\textbf{Forward Dynamics Model learned with Gaussian Process.} As a learning baseline, we use Gaussian Process regression, which is commonly used in motion planning and machine learning~\cite{Gaussian_process_ML}. Using $\mathcal{D}_{\rm sim}$, we train a Gaussian Process regressor to predict the cable endpoint location  $(\hat{x}, \hat{y})$ given input action $\ba$.


\subsection{System Ablation Comparison}
\label{ssec:system-ablation}

We compare the performance of models across $(r_0, \vmax)$ values and perform an ablation for including $\psi$ (Equation~\ref{eq:act2_wrist}).

\subsubsection{Coverage}
From simulation, we find that ${r_0 = 0.6}$ and ${\vmax = 1.8}$ provide the maximal coverage of $66.3\%$ of the reachable workspace. Comparing with other $r_0$ values with the same $\vmax$, $r_0=0.6$ tends to result in the maximum coverage, which is visualized in the top of Figure~\ref{fig:coverage}. We thus choose $r_0=0.6$.
The bottom of Figure~\ref{fig:coverage} compares the coverage with and without $\psi$ using $r_0 = 0.6$. The graph shows adding $\psi$ notably increases coverage, confirming the hypothesis in Section~\ref{ssec:action}. 

\subsubsection{Repeatability}
The physical repeatability evaluation results are shown in Figure~\ref{fig:repeat}. We observe that $\vmax = 1.2$ produces the most repeatable actions with an average standard deviation of $1.27\%$ (of cable length) across actions compared to $3.70\%$ for $\vmax = 1.8$, but offers minimal coverage $21.3\%$ of the reachable area in simulation. While $\vmax=1.8$ with $\psi$ offers the most coverage $79.7\%$, it has the least repeatability with an average standard deviation of $4.72\%$. To balance coverage and repeatability, we select $\vmax=1.5$ with $\psi$, which has a coverage of $76.9\%$ and an average standard deviation of $1.93\%$. Compared to trials using $\vmax=1.8$ with $\psi$, trials using $\vmax = 1.5$ with $\psi$ also have $17.1\%$ fewer off-table cases.


\begin{table}[t]
  \setlength\tabcolsep{5.0pt}
  \vspace{1pt}
  \centering
  \caption{
    Physical evaluation results on Cable 1, of the 2 baselines and the dynamics model $f_{\rm forw}$ trained on three different datasets. All distances are expressed as a percentage of cable length (\SI{0.62}{\meter}).
  }
  \begin{adjustbox}{width=\columnwidth,center}
    \centering
  \small
  \begin{tabular}{@{}lrrrrr}
  \toprule
  Model & Median & 1st Quartile & 3rd Quartile & Min & Max \\
  \midrule
Polar Casting & 43\%  & 14\%   & 76\% & 4\% & 106\%  \\
Gaussian Process with $\mathcal{D}_{\rm sim}$ & 29\% & 20\% & 59\% & 4\% & 159\% \\
$f_{\rm forw}$ with $\mathcal{D}_{\rm real}$ & 59\% & 43\% & 83\% & 3\% & 117\%  \\
$f_{\rm forw}$ with $\mathcal{D}_{\rm sim}$ & 29\% & 15\% & 39\% & 1\%& 113\%\\
$f_{\rm forw}$ with $\mathcal{D}_{\rm sim} + \mathcal{D}_{\rm real}$ & \textbf{22\%} &\textbf{ 9\%} & \textbf{38\%} & \textbf{0\%} & \textbf{78\% } \\
  \toprule
  \end{tabular}
  \end{adjustbox}
  \label{tab:error}
\vspace*{-5pt}
\end{table}  
\subsection{Physical Evaluation Results}
\label{ssec:physical-evaluation}
We evaluate 3 dynamics models on Cable 1, which are trained on only $\mathcal{D}_{\rm real}$, only $\mathcal{D}_{\rm sim}$, and $\mathcal{D}_{\rm sim}$ with fine-tuning on $\mathcal{D}_{\rm real}$ (i.e., $\mathcal{D}_{\rm sim} + \mathcal{D}_{\rm real}$). We find that training on $\mathcal{D}_{\rm sim}$ and fine-tuning on $\mathcal{D}_{\rm real}$ yields the best performance. Table~\ref{tab:error} summarizes physical results. The best model produces actions that have a median error distance of $22\%$ from the target. Compared to the baselines, the policy $\pi$ (Equation~\ref{eq:finv}) using the dynamics model has a lower median error distance to the target and lower IQR. Although polar casting is accurate for targets within its reachability, the limited coverage greatly increases its median distance. Training $f_{\rm forw}$ on only $\mathcal{D}_{\rm real}$ has a much higher median endpoint deviation distance, suggesting that it overfits to $\mathcal{D}_{\rm real}$ due to its small size, motivating the need to pre-train $f_{\rm forw}$ with $\mathcal{D}_{\rm sim}$. The performance improvement after fine-tuning on $\mathcal{D}_{\rm real}$ suggests real examples are useful for closing the sim-to-real gap.

\begin{table}[t]
  \setlength\tabcolsep{5.0pt}
  \centering
  \caption{
    Physical evaluation results of 3 cables (see Figure~\ref{fig:cables}) with 32 target positions for each cable, using the dynamics model $f_{\rm forw}$ trained with $\mathcal{D}_{\rm sim} + \mathcal{D}_{\rm real}$. All distances are expressed as percentages of cable length. 
  }
  \small
  \begin{adjustbox}{width=\columnwidth,center}
    \centering
  \begin{tabular}{@{}lrrrrr}
  \toprule
  Cable      & Median & 1st Quartile & 3rd Quartile  & Min & Max   \\
  \midrule
  Cable 1    & 22\% & 9\% & 38\%& 0\% &78\%   \\
  Cable 2    & 24\% & 14\% & 35\%& 1\% &77\%   \\
  Cable 3   & 34\% & 24\% & 50\%& 6\% & 89\% \\
  \toprule
  \end{tabular}
  \end{adjustbox}
  \vspace{-5pt}
  \label{tab:results}
\end{table}  


We evaluate the performance for all 32 target positions for all cables using $f_{\rm forw}$ trained with $\mathcal{D}_{\rm sim} + \mathcal{D}_{\rm real}$, and repeat each output action 5 times per target.
The 95\% confidence ellipses in Figure~\ref{fig:repeat_multi_cable} show the underlying dynamic noise of the system and the distance between the target and the center of the ellipse shown by the dashed line represents the model error. Results are summarized in Table \ref{tab:results}. We attribute the error to two sources: simulator error from examples in $\mathcal{D}_{\rm sim}$ that do not represent real cable behavior, and learning error from the forward dynamic model. Table~\ref{tab:tuning} shows the simulator tuning error, but while Cable 3 exhibits the smallest sim-to-real gap, its physical performance is the worst among all 3. We conjecture that the circular magnet at the end of Cable 3 increases stochasticity. In experiments, the magnet frequently rotates after the rest of the cable stabilizes, moving the endpoint further from the target. In future work, we will investigate whether using a different optimization strategy for tuning will mitigate this problem. While we model the extra mass at the endpoint, we may consider modeling irregular shapes to reduce the sim-to-real gap. 


\section{Conclusion and Future Work}\label{sec:conclusion}

In this paper, we present a self-supervised learning procedure for learning dynamic planar manipulation of 3 free-end cables. Experiments suggest that tuning a simulator to match the physics of real cables, then training models from data generated in simulation, results in promising physical manipulation to get cable endpoints towards desired targets.

The self-supervised fine-tuning process relies on executing hundreds of trajectories in the physical setup, which may be infeasible in other dynamic manipulation domains. Hence, we will explore dynamics randomization~\cite{dynamics_randomization_2018}. The presented method also relies on learning a separate set of simulation parameters for each cable, motivating meta-learning~\cite{maml} approaches to rapidly adapt to other cables.

\section*{Acknowledgments}

{\footnotesize
This research was performed at the AUTOLAB at UC Berkeley in affiliation with the Berkeley AI Research (BAIR) Lab, the CITRIS ``People and Robots'' (CPAR) Initiative, and the Real-Time Intelligent Secure Execution (RISE) Lab. The authors were supported in part by donations from Toyota Research Institute and by equipment grants from PhotoNeo, NVidia, and Intuitive Surgical. Daniel Seita is supported by a Graduate Fellowship for STEM Diversity (GFSD). We thank our colleagues Ashwin Balakrishna, Ellen Novoseller, and Brijen Thananjeyan.
}

\bibliographystyle{IEEEtranS}
\bibliography{example}
\normalsize
\cleardoublepage
\appendices

\end{document}